\documentclass[letterpaper, 10 pt, conference]{ieeeconf}  % Comment this line out if you need a4paper

\IEEEoverridecommandlockouts                              % This command is only needed if 
\usepackage{graphics} % for pdf, bitmapped graphics files
\usepackage{graphicx}
\usepackage{acronym}
\usepackage{amsmath}
\usepackage{hyperref}

\title{\LARGE \bf
Elevation Mapping for Locomotion and Navigation using GPU
}

\author{Takahiro Miki, Lorenz Wellhausen, Ruben Grandia, Fabian Jenelten, Timon Homberger, Marco Hutter
\thanks{This project has received funding from the European Union’s Horizon 2020 research and innovation programme under grant agreement No 780883.}% <-this % stops a space
\thanks{This research was supported by the Swiss National Science Foundation (SNSF) as part of project No.188596.}% <-this % stops a space
\thanks{This project has received funding from the European Union’s Horizon 2020 research and innovation programme under grant agreement No 101016970. 
}% <-this % stops a space
\thanks{All authors are with Robotic Systems Lab, ETH Zurich}%
}

\newacro{ROS}{Robot Operating System}
\newacro{CPU}{Central Processing Unit}
\newacro{GPU}{Graphics Processing Unit}
\newacro{CNN}{Convolutional Neural Network}
\newacro{CUDA}{Compute Unified Device Architecture}
\newacro{UAV}{Unmanned Aerial Vehicle}
\newacro{IMU}{Inertial Measurement Unit}
\newacro{SLAM}{Simultaneous Localization and Mapping}
\newacro{DARPA}{Defense Advanced Research Projects Agency}

\begin{document}

\maketitle
\thispagestyle{empty}
\pagestyle{empty}

%%%%%%%%%%%%%%%%%%%%%%%%%%%%%%%%%%%%%%%%%%%%%%%%%%%%%%%%%%%%%%%%%%%%%%%%%%%%%%%%
\begin{abstract}
Perceiving the surrounding environment is crucial for autonomous mobile robots. An elevation map provides a memory-efficient and simple yet powerful geometric representation for ground robots.
The robots can use this information for navigation in an unknown environment or perceptive locomotion control over rough terrain.
Depending on the application, various post processing steps may be incorporated, such as smoothing, inpainting or plane segmentation.
In this work, we present an elevation mapping pipeline leveraging GPU for fast and efficient processing with additional features both for navigation and locomotion. 
We demonstrated our mapping framework through extensive hardware experiments. Our mapping software was successfully deployed for underground exploration during DARPA Subterranean Challenge and for various experiments of quadrupedal locomotion.
\end{abstract}

%%%%%%%%%%%%%%%%%%%%%%%%%%%%%%%%%%%%%%%%%%%%%%%%%%%%%%%%%%%%%%%%%%%%%%%%%%%%%%%%
\section{INTRODUCTION}
For mobile robots that operate autonomously in an unknown environment, it is crucial to use their onboard sensors to perceive the surroundings.
For planning their movements, robots often create a map to fuse the sensor measurements over time since the sensor coverage is limited.
An occupancy grid can be used for indoor navigation on flat ground with obstacles~\cite{occupancy_grid,thrun2002probabilistic}. This two-dimensional map shows whether each cell is occupied or not, thus robots can use it to plan their path to the target point while avoiding obstacles for example.
However, this map is not suitable for environments with rough terrain with inclinations or steps. 
To solve this problem, a 2.5D elevation map is often used which can hold the height of the ground for each cell~\cite{herbert1989terrain,kolter2009stereo,fankhauser2014robot,fankhauser2018probabilistic}.
In this map, a 2D grid stores the height information of each cell to represent the terrain geometry.
To further extend the common 2.5D representation to multi-level structures such as a bridge, Triebel et al. proposed to construct a multi-level surface map~\cite{rudolph2006multi}.
Some methods use voxels to represent spatial information in 3D. Occupancy probabilities of voxels are updated based on sensor measurements~\cite{hornung2013octomap, oleynikova2017voxblox}.
Along with its ability to model more general structures than a
2.5D grid, the 3D grid has a higher memory requirement
compared to a 2.5D elevation map with the same grid size.

In rough terrain navigation, many planners use a cost map calculated according to local geometric features such as the slope or surface roughness~\cite{wermelinger2016, virtual_surface, fan2021step}.
Additionally, some navigation tools are based on a foothold score, as the sum of geometric features~\cite{artplanner}.
Besides mobile robot navigation, legged robots also benefit from perceiving their surroundings to control their locomotion. An elevation map provides a representation of the geometric landscape that can be directly embedded into a control pipeline. It was used to plan trajectories and select footholds in a number of works~\cite{kolter2009stereo,havoutis2013onboard,fankhauser2018robust,jenelten2020perceptive,kim2020vision,magana2019fast}.
To accomplish both navigation and locomotion tasks, it is crucial to build an elevation map and perform post-processing in real-time.

\begin{figure}[t]
   \centering
    \includegraphics[width=\columnwidth]{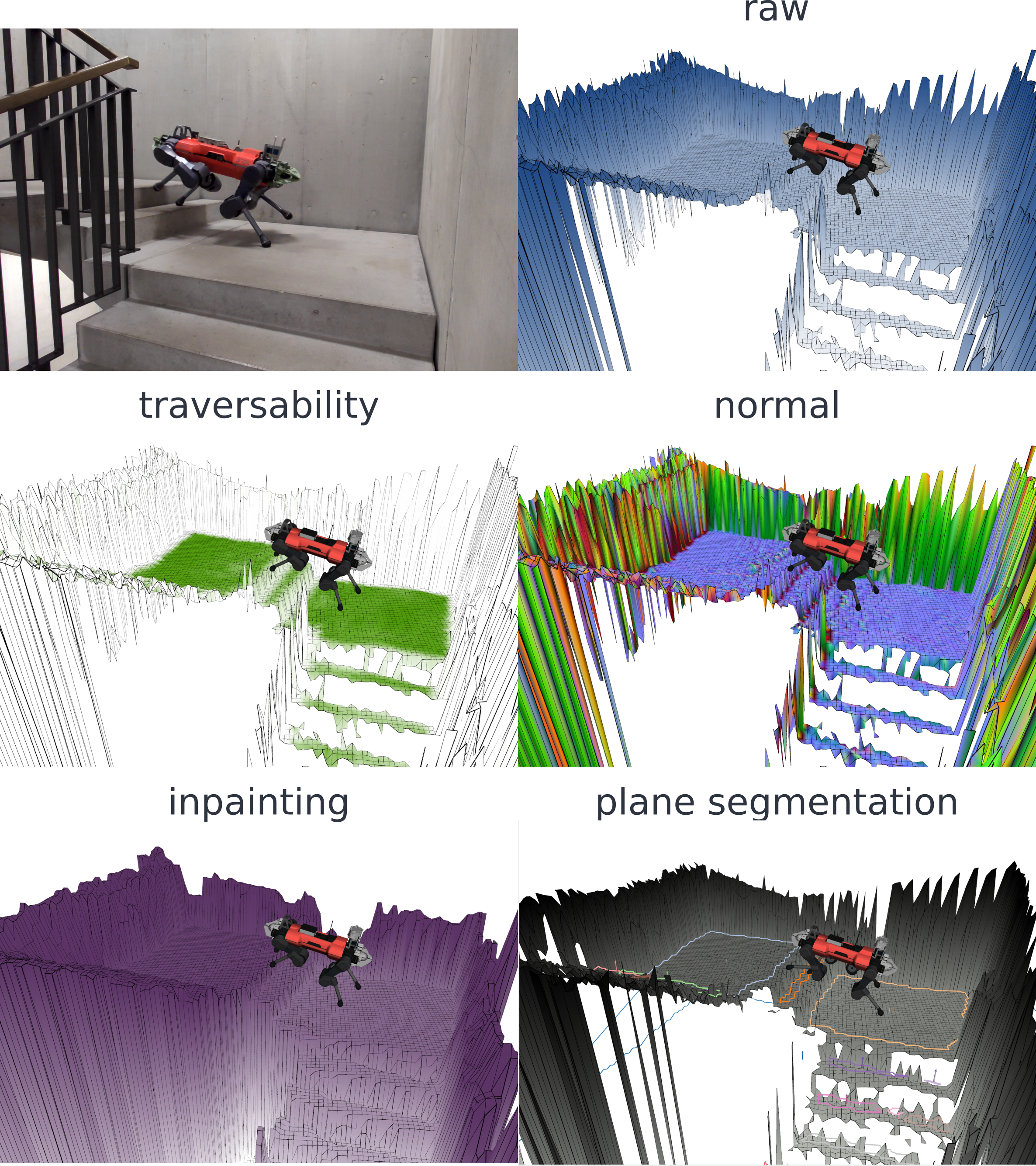}
    \caption{Proposed elevation mapping framework on \ac{GPU}. The point cloud processing, traversability estimation and normal calculation is performed on \ac{GPU}. In addition, inpainting, smoothing filters or plane segmentation features are provided for legged locomotion control.} 
    \label{fig:hiking}
\end{figure}

As an existing mapping framework, Fankhauser et al. present a probabilistic robot-centric elevation map on \ac{CPU} with \ac{ROS} integration~\cite{fankhauser2014robot,fankhauser2018probabilistic}. It transforms depth measurements, represented as point clouds, to the local frame of the robot and updates each cell's terrain height in a Kalman-Filter-like fashion. This method was used for locomotion tasks~\cite{fankhauser2018robust, buchnan2021} or navigation tasks~\cite{wermelinger2016}.
However, it is not efficient enough to process large amounts of point cloud data in real time for faster agile robot movements than the previous work~\cite{fankhauser2018robust}. This motivated our development of a terrain mapping framework with \ac{GPU} acceleration.
A similar approach was used in the recent work of Pan et al. in which a local map update was performed on \ac{GPU} and a global map was created by registering the local map~\cite{pan2021}. However, the main focus of their software is large range navigation and they incorporate down-sampling in the point cloud data. It does not consider the legged locomotion tasks that requires fine grained geometric representation.

As another approach of creating an elevation map, the voxels can be used to project a 3D map onto a 2D grid structure~\cite{fan2021step, virtual_surface, overbye2021gvom}.
There are some methods that utilize \ac{GPU} for faster point cloud processing~\cite{ohm, overbye2021gvom}.
However, to create a high-resolution terrain map, the voxel grid size needs to be small which requires a large amount of memory. In addition, the robot needs to maintain both a voxel grid and an elevation map at the same time.

This work presents an elevation mapping framework integrated with ROS using GPU for efficient point cloud registration and ray casting.
In this mapping framework, we developed and integrated multiple features that were used for legged robot navigation and locomotion.
In addition, we addressed issues we encountered during the field deployments.
For example, state estimation drift can cause artifacts in the map and to address this, we introduce height drift compensation.
Also, an overhanging obstacle near the robot will appear as a wall and prevents a planner to plan below it.
We introduced a raycasting method for visibility clean up and an exclusion area to remove virtual artifacts when the robot gets close to them.
Furthermore, we integrate a learning-based traversability filter~\cite{artplanner}, which gives a traversability value based on the local geometry and enhanced the elevation map through improvements such as \textit{upper bound} calculation or overlap clearance for navigation tasks.
For locomotion, we integrated various smoothing filters used in~\cite{jenelten2020perceptive, jenelten2021TAMOLS} and plane segmentation used in~\cite{grandia2022perceptive}.

We demonstrate our framework through extensive experiments.
Our implementation supported legged locomotion and navigation research and formed the basis for perceiving surrounding terrains used by learning-based controllers~\cite{wild_anymal, rudin2022learning, ma2022combining}, model-based controllers~\cite{jenelten2020perceptive, jenelten2021TAMOLS, grandia2022perceptive}, or others such as navigation~\cite{artplanner, yangbowen2021} or learning occlusion filling~\cite{stoelzle2022reconstructing}.
In addition, the proposed elevation mapping solution was successfully used for underground exploration missions during \ac{DARPA} Subterranean Challenge~\cite{subt} where team CERBERUS~\cite{cerberus, tranzatto2021cerberus} deployed our mapping on four quadrupedal robots, ANYmal~\cite{hutter2016anymal}. 

Our contribution is as follows,
\begin{itemize}
    \item Developing a GPU-based elevation mapping implementation, supporting a variety of different filters and other features for legged locomotion and mobile robot navigation.
    \item Validating the real-world applicability through extensive experiments.
    \item Open sourcing elevation mapping software. \footnote{\url{https://github.com/leggedrobotics/elevation_mapping_cupy}}
\end{itemize}

\section{Methods}
In this section, we describe the methods used to construct the elevation map on \ac{GPU} and the additional features used for locomotion and navigation.

We first describe the overview of our pipeline (Section \ref{sec:overview}) and show the core methodology to update the map (Section \ref{sec:details},\ref{sec:update}).
Then, we introduce features we developed to improve the map quality (Section \ref{sec:drift}, \ref{sec:visibility}, \ref{sec:overlap}). At last, we summarize the post-processing features used for navigation and legged locomotion (Section \ref{sec:traversability}, \ref{sec:upper}, \ref{sec:locomotion}).

\subsection{Overview}
\label{sec:overview}
\begin{figure}
   \centering
    \includegraphics[width=\columnwidth]{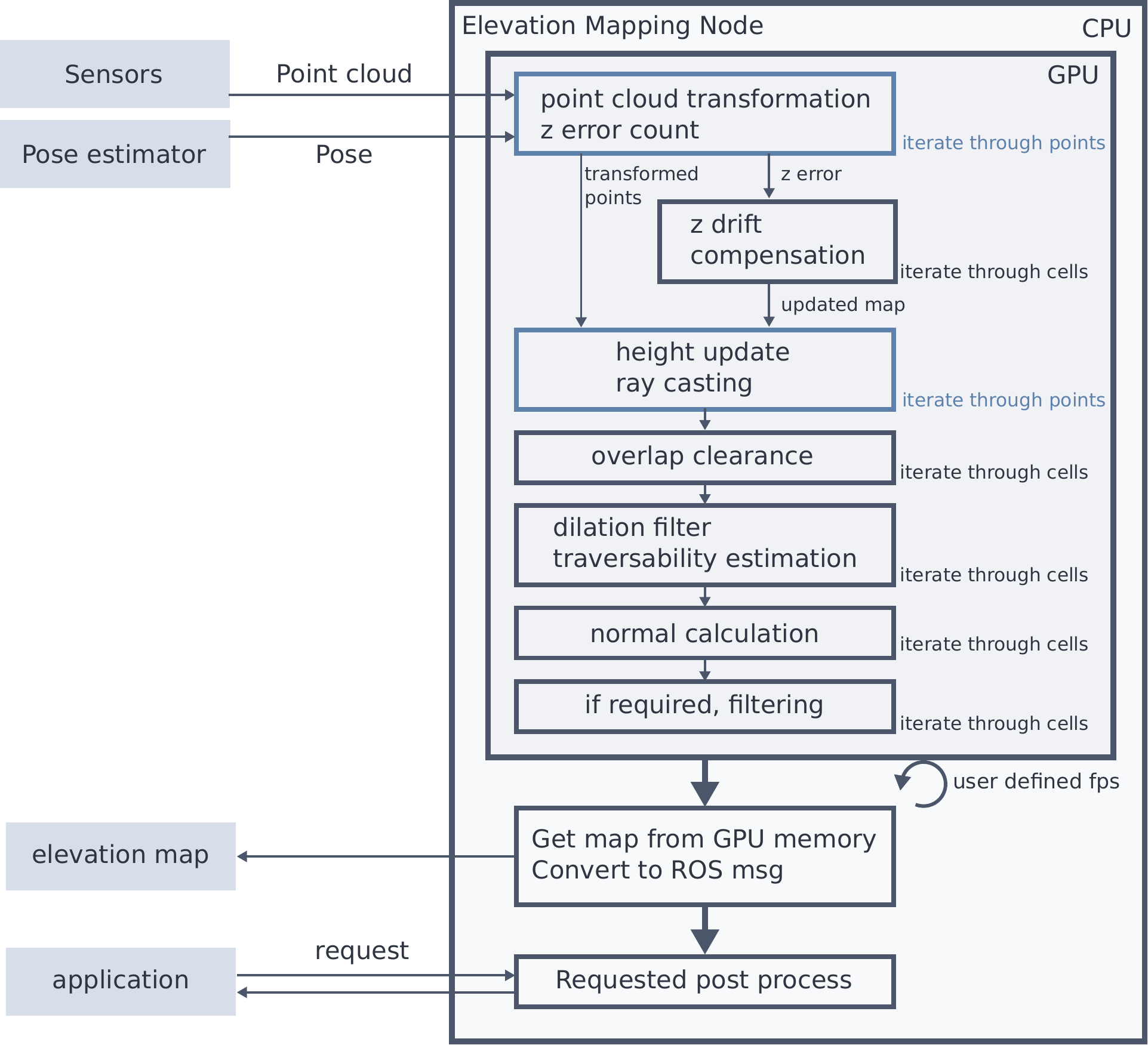}
    \caption{Overview of processing. The point cloud data from the sensor and the pose information from the pose estimator are inputs to the GPU.
    After transforming the point cloud and calculating the z drift errors, the current map estimation is adjusted to match the latest sensor measurement. Then the main height map update and ray casting are performed.
    As the map is updated, various filters are applied.
    The map data is transferred to CPU memory with a user-defined frequency to publish as a ROS message to reduce unnecessary data transmission.
}
    \label{fig:overview}
\end{figure}

Fig. \ref{fig:overview} shows the overview of our software.
The input is the point cloud collected from depth sensors and the robot's estimated pose.
The latter can be provided by a \ac{SLAM} based pose estimation or odometry system.

First, this data is transferred to \ac{GPU} memory.
Then, the points are transformed to the user specified map frame using the \ac{ROS} tf library. At the same time, height drift error is calculated and stored for the downstream process. Based on this height error, the map is shifted so that the map matches the latest sensor measurement.
Following the adjustment of the current map to the latest measurements, the height is updated through each point measurement iteration. 
Within the same iteration, we perform ray-casting to remove the penetrated objects and update the upper-bound layer. 
Then, the operations which iterate through each cell in the map are performed (overlap clearance, traversability estimation, normal calculation, and additional filtering if required).

The map is published in a user-defined frequency. With this frequency, the map data is transferred to \ac{CPU} memory and published as a GridMap~\cite{Fankhauser2016GridMapLibrary} message to reduce unnecessary data transfer between \ac{GPU} and \ac{CPU}.
Based on the request, the post-processing on \ac{CPU} is calculated and returned to the application node.

\subsection{Implementation details}
\label{sec:details}
We use \ac{ROS}~\cite{quigley2009ros} for inter-process communication and Cupy\cite{nishino2017cupy} to implement parallel computing on \ac{GPU}. We used Cupy because of its simple API in Python and the ability to define custom \ac{CUDA} Kernels. The Python interface allows using deep learning frameworks such as PyTorch~\cite{NEURIPS2019_9015} with the same \ac{GPU} memory.
We used roscpp instead of rospy mainly due to the slow serialization of \ac{ROS} messages on rospy.

\subsection{Height cell update}
\label{sec:update}
The height in each cell is updated through a Kalman filter formulation as done in \cite{fankhauser2018probabilistic}. We iterate through each point in parallel using custom \ac{CUDA} kernels.
\begin{equation}
    h = \frac{\sigma_p^2 h + \sigma_m^2 p_z}{\sigma_m^2 + \sigma_p^2},
    \label{eq:height_update}
\end{equation} where h is the estimated height of the cell, $\sigma_p^2$ is the variance of the point calculated from a sensor's noise model, and $p_z$ is the height measurement of the point. $\sigma_m^2$ is the estimated variance of the cell, initialized to a large number in the beginning.
As the noise model, we used, $\sigma_p^2 = \alpha_d d^2$, where $d$ is the distance of the point from the sensor and $\alpha_d$ is a user-defined parameter, simplified the model of Nguyen et al where the sensor noise variance increases quadratically with the distance~\cite{nguyen2012modeling}. 

Also, the variance of each cell is updated as below.
\begin{equation}
    \sigma_m^2 = \frac{\sigma_m^2 \sigma_p^2}{\sigma_m^2 + \sigma_p^2}
\end{equation}
We also add a constant time variance, $\sigma_t^2$ at a constant rate to increase the variance if the cell is not updated.

\begin{figure}
   \centering
    \includegraphics[width=1.0\columnwidth]{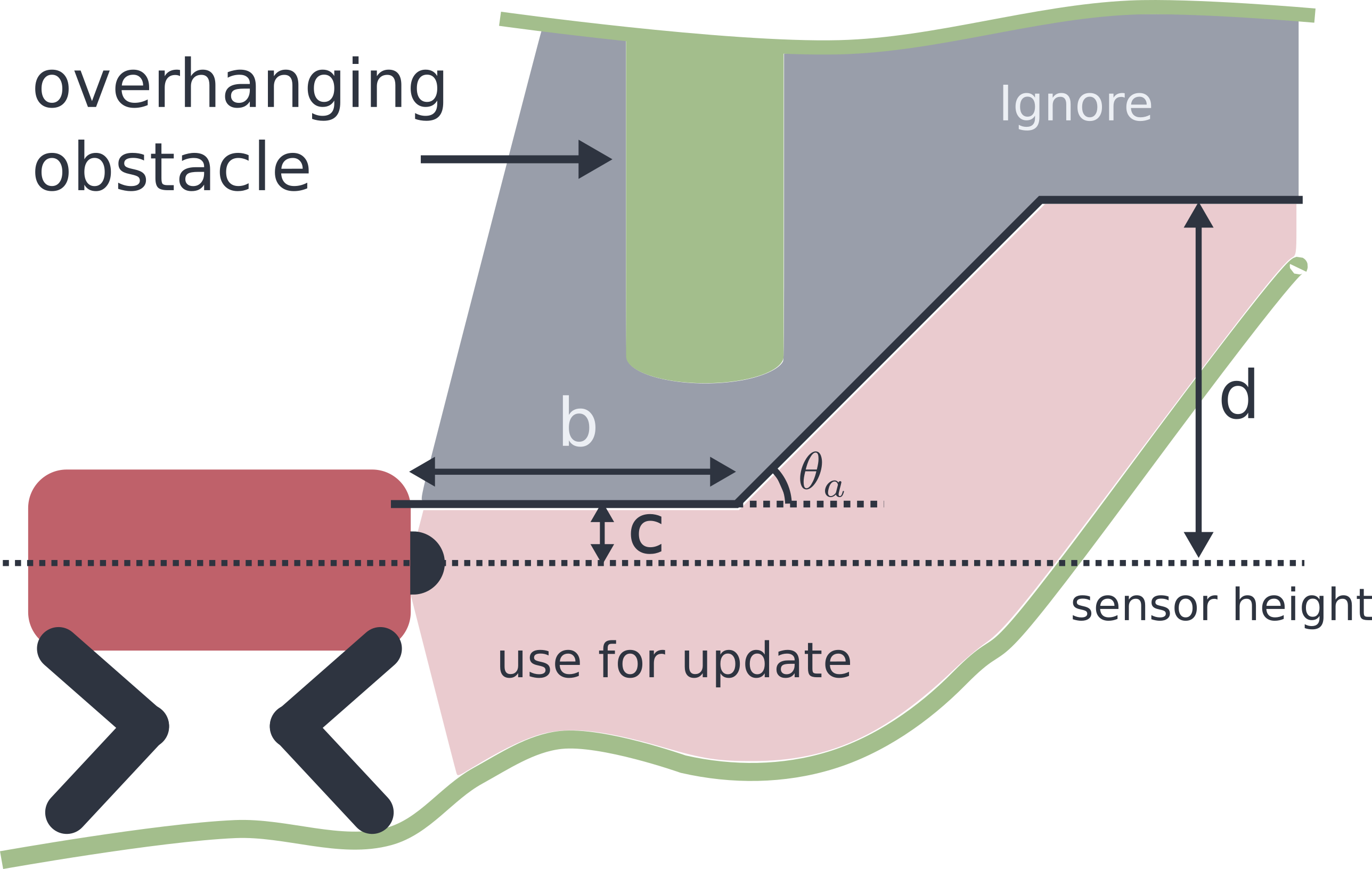}
    \caption{Exclusion area for point measurements. To avoid creating the ceiling or close overhanging obstacle as an artifact, we introduce an exclusion area defined by parameters $\theta_a, b, c, d$.} 
    \label{fig:ramp}
\end{figure}

We extend the formulation of \cite{fankhauser2018probabilistic} with additional features to generalize elevation mapping to scenarios including overhanging structures and walls.
To remove outliers, we have an outlier check based on the Mahalanobis distance as also done in the previous work~\cite{fankhauser2018probabilistic}.
Here, we reject points that have a larger height difference than a certain Mahalanobis distance.
In addition, to reject ceilings or overhanging objects while simultaneously capturing sloped terrain in the map, we introduced an exclusion area using ramp parameters as shown in Fig. \ref{fig:ramp}.
$\theta_a$ defines an angle of the exclusion area line. $b, c$ and $d$ determine the offset and maximum height as shown in Fig. \ref{fig:ramp}.
By ignoring the points above the shown lines, the map correctly expresses the small opening near the robot required for navigating through.
Furthermore, there is an issue of vertical edges or walls if we naively apply the equation \ref{eq:height_update}. The resulting height becomes lower than the actual height because there are multiple measurements of different heights on the same cell which are effectively averaged.
To resolve this, we count the number of points that fall within the same cell, and if it is higher than a threshold, we ignore the points lower than the currently estimated height. Since this happens before the outlier detection, which increases the variance of the cell, this can make the edge sharper.

\subsection{Height drift compensation}
\label{sec:drift}
\begin{figure}
   \centering
    \includegraphics[width=\columnwidth]{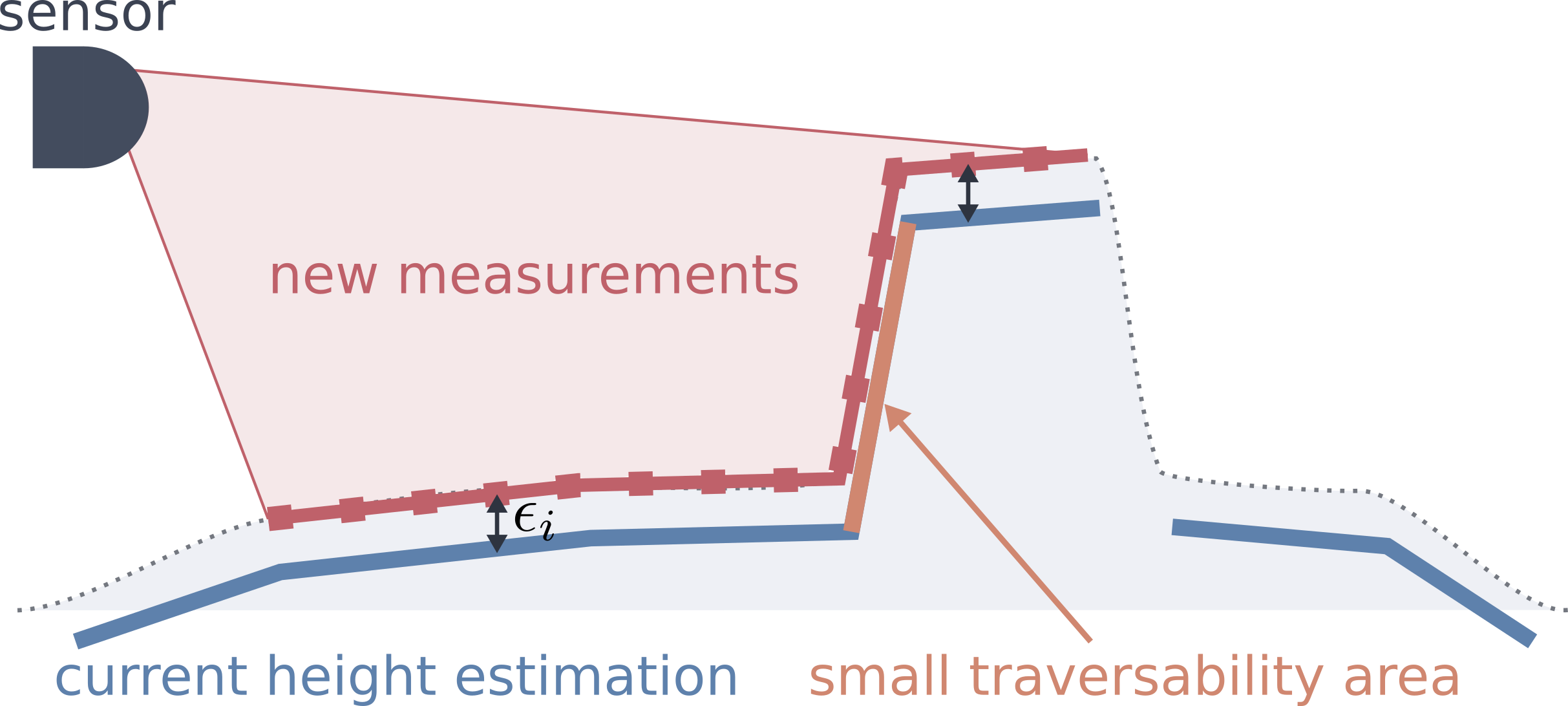}
    \caption{Height drift compensation. To reduce the effect of the state estimation drift, we perform a height drift compensation using the new measurements. The height error between the new measurements and the current height estimation is calculated while excluding the area with low traversability area to prevent miscalculation using points with various heights at the same cell. At last, the whole map is adapted to the new measurements.}
    \label{fig:drift}
\end{figure}

Since it is difficult to perfectly estimate the state onboard, the estimated sensor position with respect to the map often drifts.
Height drift is the most influencing factor since it causes artifacts in the map.
Therefore, a simple drift compensation method is developed to reduce this issue by calculating the error between the current sensor measurement and the map, and adjusting the map accordingly.
As seen in Fig. \ref{fig:drift}, the error $\epsilon_i$ is calculated for each point $p_i$.
To avoid projection errors in rough or steep areas, we only use relatively flat structure to calculate the error by thresholding the cell's traversability value (see \ref{sec:traversability}).
Once all points have been processed, the average error $\frac{1}{n}\sum_i^n \epsilon_i$ is added to the elevation layer in the map, where $n$ is the number of points used to calculate the error.

\subsection{Visibility cleanup}
\label{sec:visibility}
In case there are dynamic obstacles, the old obstacles remain in the map until the estimated variance becomes large enough to pass the outlier rejection.
Visibility cleanup is performed to remove this remaining estimate by performing ray casting.
For each point, a ray from the sensor origin to the measurement point is cast and we iterate through this ray with a certain step size.

\begin{figure}
   \centering
    \includegraphics[width=\columnwidth]{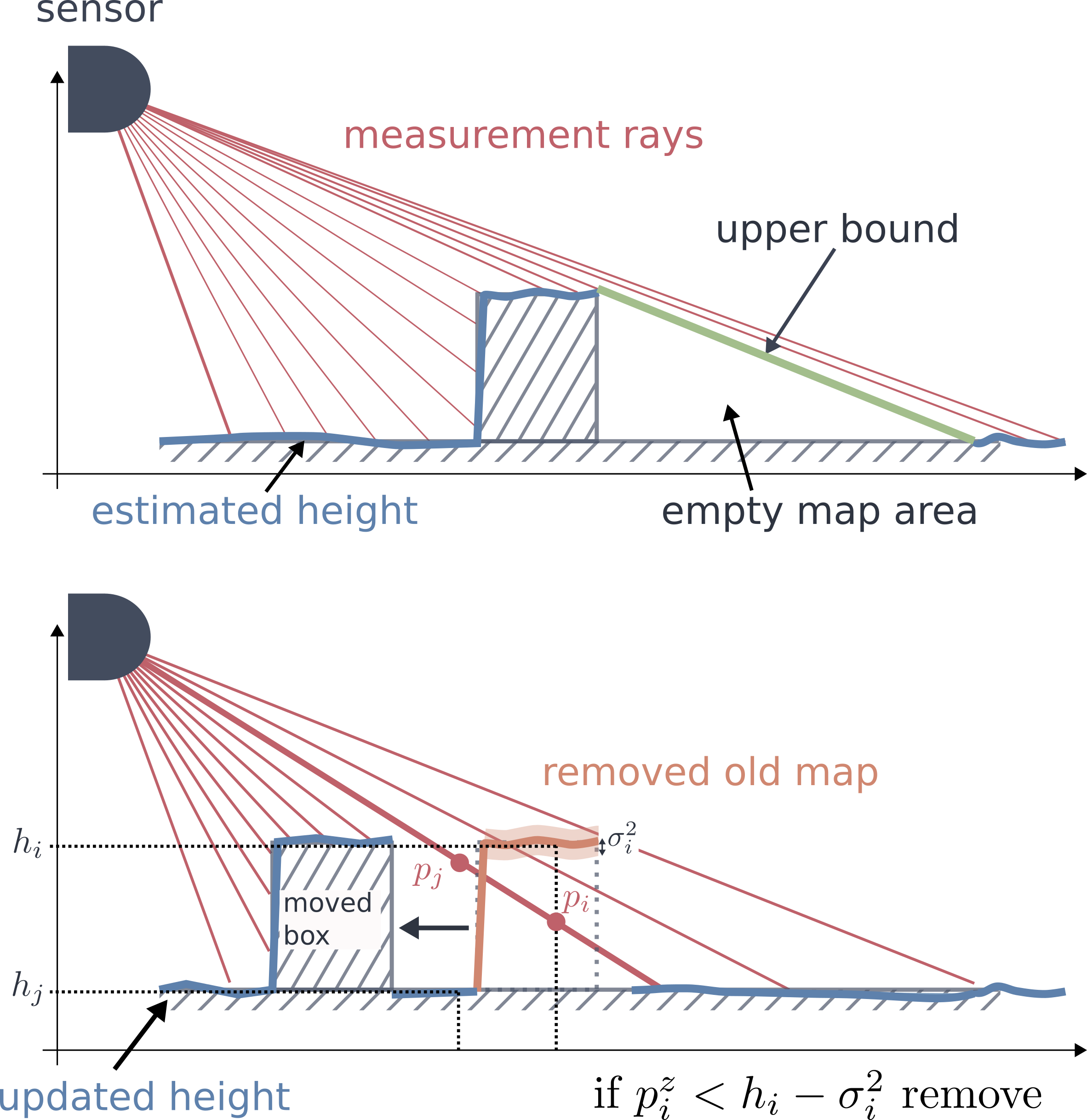}
    \caption{Visibility cleanup and \textit{upper bound} calculation using ray casting.
    The above figure shows the \textit{upper bound} region. Based on the measured ray, the \textit{upper bound} value in the area without a valid estimation is updated. The bottom figure shows the visibility cleanup. If the ray goes through the old map, that cell is removed to update the map with dynamic obstacles faster.
    } 
    \label{fig:raycasting}
\end{figure}

To clear dynamic obstacles, we check if a ray went through it by comparing the ray's height and the cell's estimated height.
If a point on the ray is below the estimated height minus its variance $\sigma_i^2$, this cell is considered to be penetrated by the ray, and thus, the height value is removed:
\begin{equation}
    \text{if } p_i^z < h_i - \sigma_i, \text{ remove}.
\end{equation}, where $p_i^z$ represents a height of a point along a ray and $h_i$ represents the height of the cell that the point $p_i$ belongs to.
As seen in Fig. \ref{fig:raycasting}, in case of the point $p_j$, $p_j^z$ is higher than $h_j$ so this cell is not removed.

The removal-law can cause jittering in between two consecutive updates, e.g. when the ray hits the obstacle at a small angle.
To prevent this, we additionally check the last updated time and surface normal.
The cell is removed only if the cell has not been updated for a specific time and fulfills the normal condition,
\begin{equation}
    |\mathbf{r} \cdot \mathbf{n} | > \alpha_n,
\end{equation} where $\mathbf{r}$ represents a unit vector parallel to the ray and $\mathbf{n}$ represents the estimated surface normal of the cell. $\alpha_n$ is a user defined threshold.

\subsection{Overlap clearance}
\label{sec:overlap}
When the robot navigates through multi floor environments, the 2.5D elevation map faces an issue of remaining old height estimates from the previous upper or lower floor. To handle this, we introduced an overlapping cleaner: It clears the height value whose difference from the robot's height is larger than a threshold if it is close to the robot.
This simple method works well in multi floor environment where the robot goes up or down stairs.

\subsection{Learning based traversability filter}
\label{sec:traversability}
An important benefit of performing elevation mapping on \ac{GPU} is that data is already loaded into memory and we can therefore perform efficient terrain analysis using neural networks without any processing overhead caused by data transfer between CPU and GPU.
Specifically, we deploy a simple \ac{CNN} model trained to output traversability values for robot navigation~\cite{artplanner}.
It is implemented in PyTorch~\cite{NEURIPS2019_9015} and is light enough to run at full map update rate.

\subsection{Height upper bound layer}
\label{sec:upper}
When we perform the visibility cleanup, we also use the ray casting steps to compute a maximal terrain height for unobserved map cells, similar to the idea of virtual surfaces~\cite{virtual_surface}, which we store in a map layer dubbed \textit{upper bound}.
If a ray passes through a cell, we know that the ground height cannot be larger than the ray height, since we would have otherwise observed the ground.
This is useful in case of obstacle occlusion, as holes in the map need to somehow be interpreted by the downstream modules.
A navigation planner, for example, would typically either optimistically fill the holes with an image inpainting algorithm\cite{yangbowen2021} or simply consider them untraversable.
The \textit{upper bound} layer helps distinguish between safe patches caused by obstacle occlusions (which typically exhibit small inclinations in the \textit{upper bound}) and unsafe larger drops which cause steeper ray angles resulting in larger inclinations.

\subsection{Features for legged locomotion}
\label{sec:locomotion}
Elevation maps can be used not only for navigation, but also for foothold planning for legged robots.
Some additional post processing is often required to use in optimization based locomotion controllers.
We provide optional features that are particularly useful for such applications.

\subsubsection{Minimum filter}
The completeness of the elevation map is crucial for reliable and robust motion planning in model based controllers. We provide an implementation of the in-painting filter presented in~\cite{jenelten2021TAMOLS}, which replaces empty grid cells with the minimum found along the occlusion border.

\subsubsection{Smoothing filter}
Gaussian and box-blur filters have been used in~\cite{jenelten2020perceptive, jenelten2021TAMOLS} for generating smooth representations of the terrain. Median filtering has been reported as being more effective for artifact rejection~\cite{jenelten2021TAMOLS}. We provide combinations of several smoothing filters that can be found in the \texttt{opencv} library~\cite{opencv_library}, including Gaussian, box-blur and median filter.

\subsubsection{Plane segmentation}
In an optimization-based approach to perceptive locomotion, the terrain is often represented by a set of primitive geometric shapes to optimize over. In favour of low computation time, it is beneficial to globally extract these features before the optimization is started. We provide algorithms to segment the elevation map into continuous planar regions. The outer boundary and potential holes contained inside the region are returned as polygons. This preprocessing is used in the whole-body MPC approach presented in~\cite{grandia2022perceptive}.

\section{Experiments and results}
To evaluate our mapping framework, we first compared our mapping framework with existing open source software~\cite{fankhauser2018probabilistic} as a baseline.
Then, we validate the usability of our mapping framework through legged locomotion and navigation experiments using the ANYmal platform~\cite{anymal}.
The robot used in our experimens is equipped with two Robosense Bpearl~\cite{bpearl} LiDARs or four Intel Realsense~\cite{realsense}.

\subsection{Baseline comparison}
\subsubsection{Feature evaluation}
\begin{figure}
   \centering
    \includegraphics[width=\columnwidth]{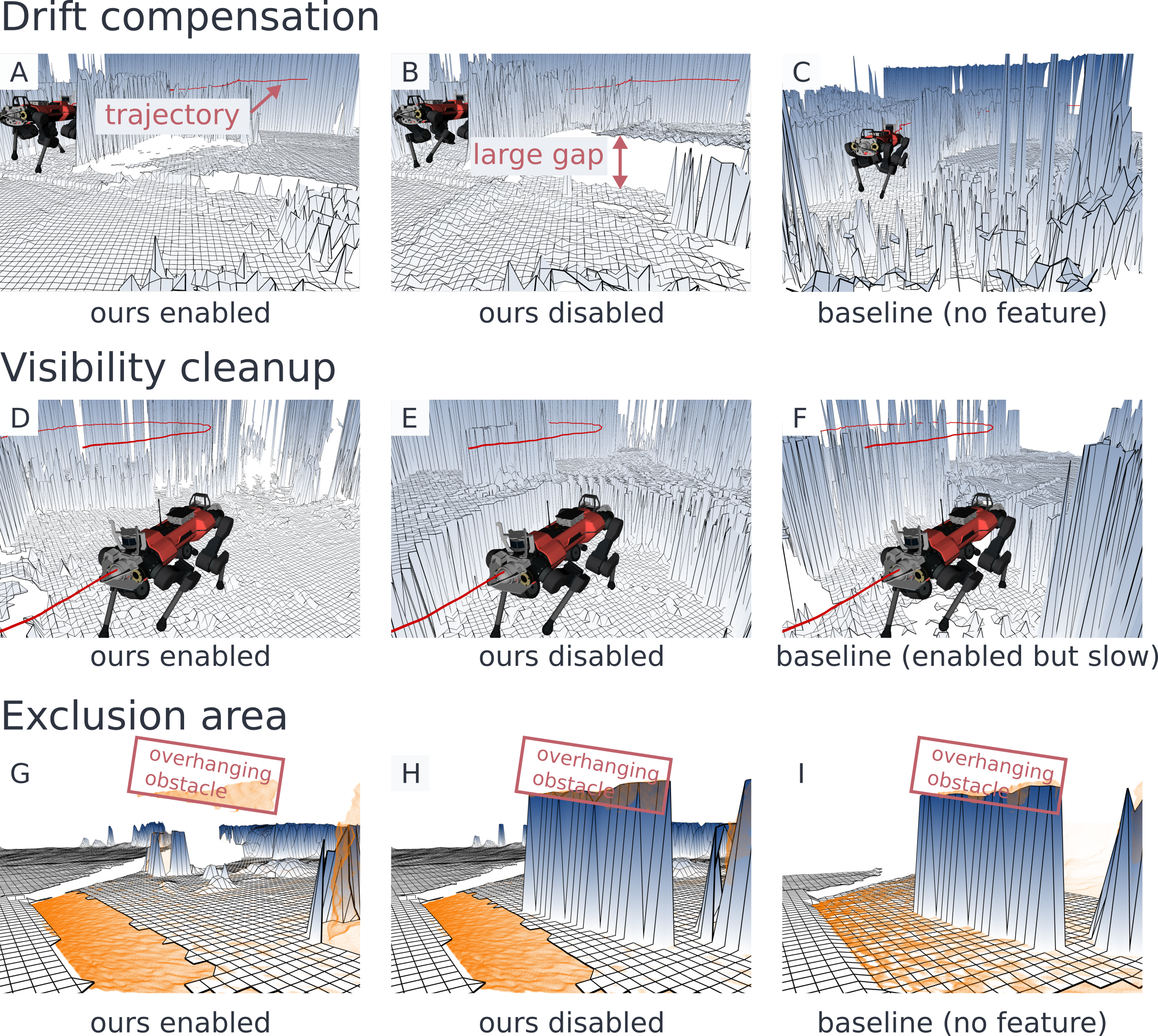}
    \caption{Feature comparison. We evaluated the effectiveness of each features by comparing the map quality with feature disabled setting. In addition, we compared with the baseline method~\cite{fankhauser2018probabilistic}.
    Drift compensation: when the odometry has a large height drift, it causes a large gap as seen in B. Our drift compensation filter can reduce this by matching the current map to the new measurement.
    Visbility cleanup: Our visibility cleanup is performed on every point cloud measurement instead of slower rate as done in the baseline. This results in much cleaner map as seen in D.
    Exclusion area: We placed an overhanging obstacle in front of the robot. As seen in G, our method could exclude this while the baseline method represented as an obstacle.
    } 
    \label{fig:feature_comparison}
\end{figure}

Here, we first evaluate the effectiveness of our features by comparing the quality of the map using the same data collected by the ANYmal (Fig. \ref{fig:feature_comparison}).
In the first two row (Fig. \ref{fig:feature_comparison} A-F), the robot was equipped with two Bpearls and the state estimation had a large height bias as shown by the trajectory of the robot (shown in red line). 
With our drift compensation, the gap between the old map and the latest map updated by the new sensor measurement was smaller than the one without this feature (Fig. \ref{fig:feature_comparison} A-C). The map of the baseline creates a large gap and also some artifacts. This artifacts was due to the slow visibility cleanup rate as seen in the following experiment.

In the second row (Fig. \ref{fig:feature_comparison} D-F), we evaluate the visibility cleanup feature using the same data as Fig. \ref{fig:feature_comparison} A-C.
In this setting, we disabled the drift compensation and compared the map quality.
As seen in Fig. \ref{fig:feature_comparison}D, our map with visibility cleanup feature did not have the wall which is visible in Fig. \ref{fig:feature_comparison}E. This is because the new measurement could remove the old map through ray casting.
The baseline framework has the visibility cleanup feature, however, the update is performed in a slower rate than the point cloud measurements to reduce the computational load. As seen in Fig. \ref{fig:feature_comparison}F this slow update rate could not clear all artifacts.

Lastly we compared the map with an overhanging obstacle in front of the robot.
As seen in Fig. \ref{fig:feature_comparison}G-I, our method could exclude the obstacle while the baseline method represented it as a wall. 

\subsubsection{Processing time comparison}
\label{sec: comparison}
We compared the processing time of our elevation mapping pipeline on \ac{GPU} with the baseline.
We measured the calculation time of processing the point cloud on two devices,
a desktop PC (Ryzen9 3950x, NVIDIA RTX 2080Ti), as well as a Jetson Xavier that is integrated on the robot.
To measure the calculation time with different number of points, we randomly sampled from a point cloud data and republished to the mapping pipeline.
We used the map size of $10\times10$ m with 4cm resolution and 
the calculation time was measured on \ac{GPU} for our pipeline and on \ac{CPU} for the baseline.
In our mapping method, the processing includes all features as shown in Fig. \ref{fig:overview}: point cloud transformation, noise handling, height error count, drift compensation, height update, ray casting, traversability estimation and normal calculation.
In the case of the baseline method, it includes the point cloud transformation, noise handling and updating each height but does not include other features such as visibility cleanup or filtering.

\begin{figure}
   \centering
    \includegraphics[width=\columnwidth]{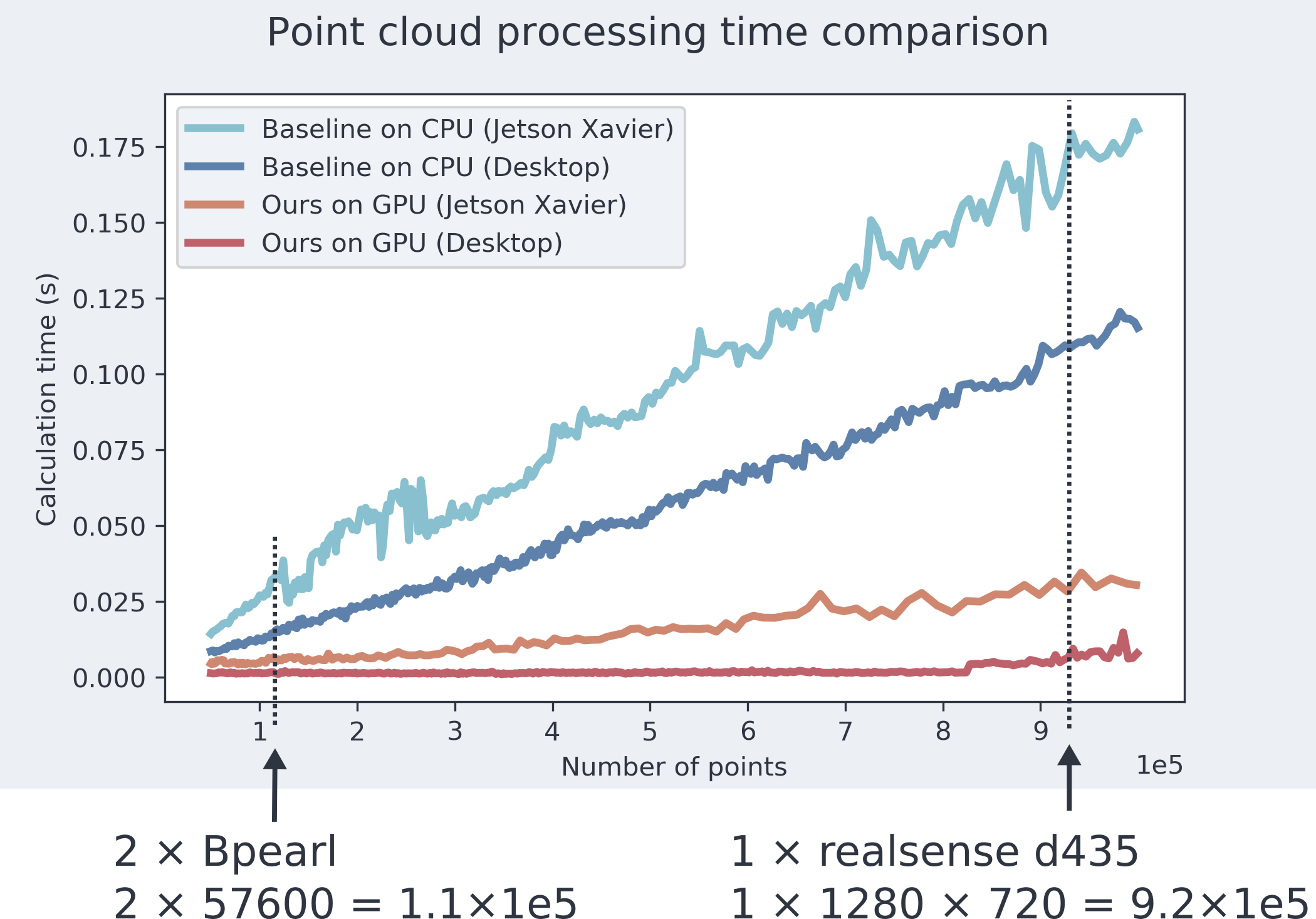}
    \caption{Point cloud processing time comparison between our method and an existing implementation on CPU~\cite{fankhauser2018probabilistic}. The calculation time increases more with the baseline method while it stays comparatively low with our method. We indicated number of points with 2xBpearl~\cite{bpearl} and 1xrealsense~\cite{realsense} on the plot. Our mapping pipeline could process the data in real-time, while the baseline method had a considerable delay on onboard PC (Jetson).} 
    \label{fig:processing_time}
\end{figure}

The result is shown in Fig. \ref{fig:processing_time}. 
Since our processing is done on \ac{GPU}, the processing time remained short even for large numbers of points.
While the baseline method's calculation time grew with a steeper rate with respect to the number of points.
This shows that our mapping can process point cloud data more efficiently than the baseline although our mapping is performing additional computations such as traversability estimation and ray casting.

\subsection{Performance analysis}
In addition, we measured the computational time per each features to analyze which component takes time.
The map size setting was the same as \ref{sec: comparison}.
The result is on TABLE \ref{tab:time}. We used Bpearl sensor data to measure the calculation time for each component on Jetson Xavier and collected 1000 measurements.
As a result, the traversability filter took the most time compared to the other components. Therefore, improving this module could improve the performance in the future.

\begin{table}[]
\caption{Calculation time per features.}
\label{tab:time}
\centering
\begin{tabular}{lrl}
number of points                 & 43017 &  \\
point transform \& z error count & 1.194 &  ms \\
drift compensation               & 0.742 &  ms \\
height update \& ray casting     & 0.648 &  ms \\
overlap clearance                & 0.003 &  ms \\
traversability                   & 4.102 &  ms \\
normal calculation               & 0.168 &  ms \\
total                            & 6.857 &  ms 
\end{tabular}
\end{table}

Additionally, to check the real-time performance on different sensor configuration, we measured the map update frequency on Jetson Xavier. This frequency shows how many times the map was updated using the sensor measurements. As seen in TABLE \ref{tab:fps}, there are three settings: Realsense filtered, Realsense raw and Bpearl.
The map setting was the same as \ref{sec: comparison}.
The \textit{Realsense filtered} setting uses a down-sampled point cloud using voxel filter implementation~\cite{Rusu_ICRA2011_PCL}.
In this setup we could achieve the update rate of almost 50 Hz, which is fast enough for locomotion.
The \textit{Realsense raw} setting uses the raw point cloud coming from the sensor. Although this point cloud contains large number of point cloud, we could update the map in 16 Hz, which is still fast enough for our use case.
The \textit{Bpearl} uses the raw point cloud coming from the Bpearl sensor. Since this sensor is sparse compared to depth image of Realsense, we could update the map in the same frequency as the sensor measurements.

\begin{table}[]
\caption{Point cloud processing frequency with different sensors.}
\centering
\begin{tabular}{lrrr}
\label{tab:fps}
Sensor             & \multicolumn{1}{l}{Number of points} & \multicolumn{1}{l}{map update Hz} & \multicolumn{1}{l}{sensor Hz} \\ \hline
Realsense filtered & 6276                                 & 49.4                               & 60                                     \\
Realsense raw      & 407040                               & 16.1                               & 60                                     \\
Bpearl             & 43074                                & 19.99                              & 20                                    
\end{tabular}
\end{table}

\subsection{Locomotion and navigation applications}
\begin{figure}
   \centering
    \includegraphics[width=\columnwidth]{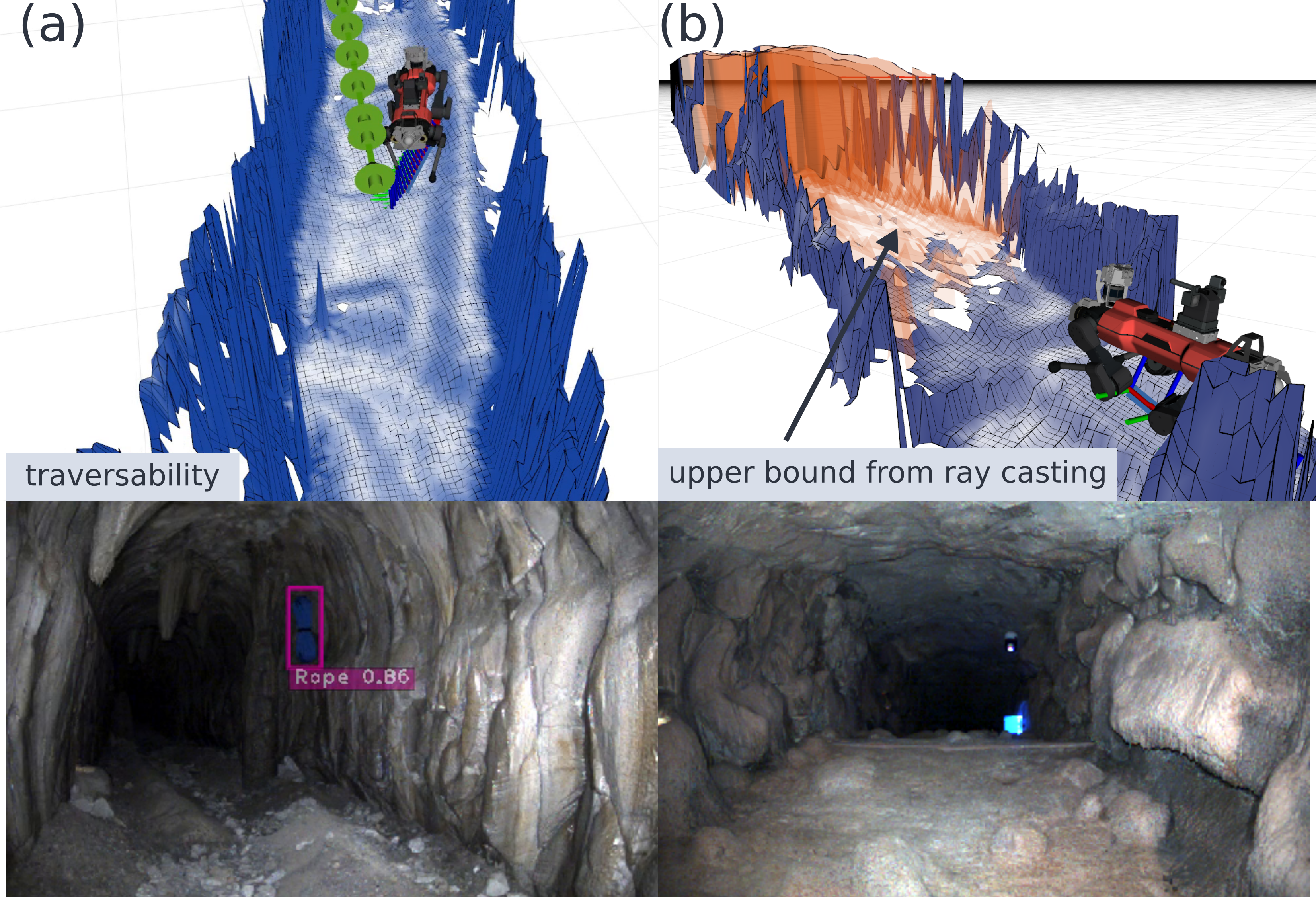}
    \caption{Elevation map used during \ac{DARPA} SubT Challenge final. The bottom images are from the on-board camera on the robot. (a) The robot navigating through a rough and narrow cave section. The traversability layer was used for local navigation. The blue color represents small traversability value and white means high traversability. (b) The robot was on a steep and narrow slope. At the end of the slope there was a flat part as shown in the bottom image. The orange map shows the upper bound layer that is calculated from the ray casting. This layer helps planning on a slope where the area are not visible due to occlusion.} 
    \label{fig:darpa}
\end{figure}

We validate the usability of our mapping framework through legged locomotion and navigation experiments using the ANYmal platform~\cite{anymal}.
We will first present the usage of the elevation map for navigation in the context of the DARPA Subterranean Challenge (Fig. \ref{fig:darpa}).
As odometry source, we rely either on kinematics based leg odometry~\cite{bloesch2013state} or a fused odometry using a time offset compensating Extended Kalman Filter~\cite{lynen13robust} that combines \ac{IMU} measurements and LiDAR \ac{SLAM}~\cite{khattak2020complementary} pose estimates.

\begin{figure*}[htbp]
   \centering
    \includegraphics[width=2\columnwidth]{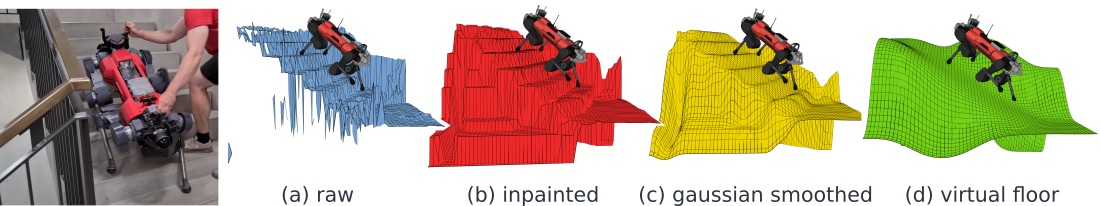}
    \caption{Different height layers used in TAMOLS~\cite{jenelten2021TAMOLS}: First, the occluded regions in the raw map are in-painted. Then, two additional layers are derived from it, gradually increasing the smoothness.} 
    \label{fig:fabian}
\end{figure*}

\subsubsection{Using traversability value for local path planning}
An important component in the autonomous mission stack is the local navigation. Our elevation map was used as a basis for local navigation to choose a feasible path and preventing the robot to walk into dangerous places such as cliffs. Given a mid range target position defined from an exploration planner~\cite{kulkarni2021autonomous}, the local planner~\cite{artplanner} plans a path using the upper bound layer and the traversability layer.

\subsubsection{Using elevation map for locomotion}
To perceive the surrounding information, the reinforcement learning-based controller~\cite{wild_anymal} sampled heights around each foot from our elevation map.
More details are in section~\ref{sec:wild_anymal}.

\subsection{Legged locomotion}
In addition to mobile robot navigation, our elevation map was used for perceptive locomotion.
We demonstrated the usability with three different controllers: A reinforcement learning based controller and two model-based controllers.
\subsubsection{Learning based locomotion}
\label{sec:wild_anymal}
The controller~\cite{wild_anymal}, which was also used during the \ac{DARPA} SubT Challenge, utilized the elevation information from our map.
One distinct feature of our mapping approach was the height drift compensation. 
When using leg odometry~\cite{bloesch2013state} as a source of sensor position estimates in the walking experiments, we observed a tendency for vertical drift, especially when the robot was significantly slipping.
The drift causes artifacts in the map that can appear as a step.
Enabling drift compensation helped the controller achieving smoother locomotion. In long missions on challenging surfaces we observed that instead relying on the fusion based reference odometry with IMU and LiDAR SLAM combined with drift compensation reduces these issues even further.
With this approach we prevent leg slippage induced odometry drift from directly affecting the elevation map quality.
\subsubsection{Model based locomotion}
Our elevation mapping framework has been used in two model based locomotion planners, utilizing the fast update rate. In~\cite{jenelten2020perceptive}, filters for computing terrain normal and standard deviation were used to compute a foothold score. The authors of~\cite{jenelten2021TAMOLS} proposed to use a virtual floor (see Fig.~\ref{fig:fabian}) as base pose reference, computed with box-blur, dilation and median filters. 
In~\cite{grandia2022perceptive}, a plane segmentation was used to define a foot hold constraint for optimization based controller.

\section{CONCLUSIONS}
In this paper, we present an elevation mapping pipeline on \ac{GPU}. %with features for navigation and locomotion.
Thanks to the \ac{GPU} acceleration, the point cloud processing is more efficient compared to similar baseline software.
In addition, several methods such as height drift compensation, upper bound calculation, or exclusion area were introduced to make the map more reliable.
We showed that our mapping methods can address issues which cause inaccurate map with the baseline framework. 
We further demonstrated the usefulness of our mapping for both navigation and legged locomotion through hardware experiments.
Our map was deployed during \ac{DARPA} Subterranean Challenge and was used for local navigation as well as perceptive locomotion.
Besides, our map supported successful application for legged locomotion over complex terrain.
The features such as smoothness filters or plane segmentation were integrated into our mapping framework, making this as a convenient tool for legged locomotion research.

%%%%%%%%%%%%%%%%%%%%%%%%%%%%%%%%%%%%%%%%%%%%%%%%%%%%%%%%%%%%%%%%%%%%%%%%%%%%%%%%

%%%%%%%%%%%%%%%%%%%%%%%%%%%%%%%%%%%%%%%%%%%%%%%%%%%%%%%%%%%%%%%%%%%%%%%%%%%%%%%%



%%%%%%%%%%%%%%%%%%%%%%%%%%%%%%%%%%%%%%%%%%%%%%%%%%%%%%%%%%%%%%%%%%%%%%%%%%%%%%%%

% References are important to the reader; therefore, each citation must be complete and correct. If at all possible, references should be commonly available publications.

% \bibliographystyle{IEEEabrv}

% \bibliographystyle{bibliography/IEEEtran}
\bibliographystyle{plain}
\bibliography{sample}

\begin{thebibliography}{10}

\bibitem{realsense}
Intel realsense (2021, april).
\newblock \url{https://www.intelrealsense.com/}.

\bibitem{bpearl}
Rs-bpearl (2021, april).
\newblock \url{https://www.robosense.ai/en/rslidar/RS-Bpearl}.

\bibitem{ohm}
Occupancy homogeneous map.
\newblock \url{https://github.com/csiro-robotics/ohm}, 2018.

\bibitem{anymal}
{ANYbotics}.
\newblock {ANYmal}.
\newblock \url{https://www.anybotics.com/anymal-autonomous-legged-robot/},
  2021.
\newblock [Online; accessed June-2021].

\bibitem{bloesch2013state}
Michael Bloesch, Marco Hutter, Mark~A Hoepflinger, Stefan Leutenegger,
  Christian Gehring, C~David Remy, and Roland Siegwart.
\newblock State estimation for legged robots-consistent fusion of leg
  kinematics and imu.
\newblock {\em Robotics}, 17:17--24, 2013.

\bibitem{opencv_library}
G.~Bradski.
\newblock {The OpenCV Library}.
\newblock {\em Dr. Dobb's Journal of Software Tools}, 2000.

\bibitem{buchnan2021}
Russell Buchanan, Lorenz Wellhausen, Marko Bjelonic, Tirthankar Bandyopadhyay,
  Navinda Kottege, and Marco Hutter.
\newblock Perceptive whole-body planning for multilegged robots in confined
  spaces.
\newblock {\em Journal of Field Robotics}, 38(1):68--84, 2021.

\bibitem{cerberus}
{Cerberus}.
\newblock Team cerberus.
\newblock \url{https://www.subt-cerberus.org/}, 2021.
\newblock [Online; accessed June-2021].

\bibitem{subt}
{DARPA}.
\newblock Darpa subterranean challenge competition rules final event.
\newblock \url{https://www.subtchallenge.com}, 2021.
\newblock [Online; accessed June-2021].

\bibitem{occupancy_grid}
A.~Elfes.
\newblock Using occupancy grids for mobile robot perception and navigation.
\newblock {\em Computer}, 22(6):46--57, 1989.

\bibitem{fan2021step}
David~D Fan, Kyohei Otsu, Yuki Kubo, Anushri Dixit, Joel Burdick, and Ali-Akbar
  Agha-Mohammadi.
\newblock Step: Stochastic traversability evaluation and planning for
  risk-aware off-road navigation.
\newblock {\em arXiv preprint arXiv:2103.02828}, 2021.

\bibitem{fankhauser2018robust}
P{\'e}ter Fankhauser, Marko Bjelonic, C~Dario Bellicoso, Takahiro Miki, and
  Marco Hutter.
\newblock Robust rough-terrain locomotion with a quadrupedal robot.
\newblock In {\em 2018 IEEE International Conference on Robotics and Automation
  (ICRA)}, pages 5761--5768. IEEE, 2018.

\bibitem{fankhauser2014robot}
P{\'e}ter Fankhauser, Michael Bloesch, Christian Gehring, Marco Hutter, and
  Roland Siegwart.
\newblock Robot-centric elevation mapping with uncertainty estimates.
\newblock In {\em Mobile Service Robotics}, pages 433--440. World Scientific,
  2014.

\bibitem{fankhauser2018probabilistic}
P{\'e}ter Fankhauser, Michael Bloesch, and Marco Hutter.
\newblock Probabilistic terrain mapping for mobile robots with uncertain
  localization.
\newblock {\em IEEE Robotics and Automation Letters}, 3(4):3019--3026, 2018.

\bibitem{Fankhauser2016GridMapLibrary}
P{\'{e}}ter Fankhauser and Marco Hutter.
\newblock {A Universal Grid Map Library: Implementation and Use Case for Rough
  Terrain Navigation}.
\newblock In Anis Koubaa, editor, {\em Robot Operating System (ROS) – The
  Complete Reference (Volume 1)}, chapter~5. Springer, 2016.

\bibitem{grandia2022perceptive}
Ruben Grandia, Fabian Jenelten, Shaohui Yang, Farbod Farshidian, and Marco
  Hutter.
\newblock Perceptive locomotion through nonlinear model predictive control.
\newblock {\em submitted to IEEE Transactions on Robotics}, 2022.

\bibitem{havoutis2013onboard}
Ioannis Havoutis, Jesus Ortiz, Stephane Bazeille, Victor Barasuol, Claudio
  Semini, and Darwin~G Caldwell.
\newblock Onboard perception-based trotting and crawling with the hydraulic
  quadruped robot ({HyQ}).
\newblock In {\em 2013 IEEE/RSJ International Conference on Intelligent Robots
  and Systems}, pages 6052--6057. IEEE, 2013.

\bibitem{herbert1989terrain}
M.~Herbert, C.~Caillas, E.~Krotkov, I.S. Kweon, and T.~Kanade.
\newblock Terrain mapping for a roving planetary explorer.
\newblock In {\em Proceedings, 1989 International Conference on Robotics and
  Automation}, pages 997--1002 vol.2, 1989.

\bibitem{virtual_surface}
Thomas Hines, Kazys Stepanas, Fletcher Talbot, Inkyu Sa, Jake Lewis, Emili
  Hernandez, Navinda Kottege, and Nicolas Hudson.
\newblock Virtual surfaces and attitude aware planning and behaviours for
  negative obstacle navigation.
\newblock {\em IEEE Robotics and Automation Letters}, 6(2):4048--4055, 2021.

\bibitem{hornung2013octomap}
Armin Hornung, Kai~M Wurm, Maren Bennewitz, Cyrill Stachniss, and Wolfram
  Burgard.
\newblock Octomap: An efficient probabilistic 3d mapping framework based on
  octrees.
\newblock {\em Autonomous robots}, 34(3):189--206, 2013.

\bibitem{hutter2016anymal}
Marco Hutter, Christian Gehring, Dominic Jud, Andreas Lauber, C~Dario
  Bellicoso, Vassilios Tsounis, Jemin Hwangbo, Karen Bodie, Peter Fankhauser,
  Michael Bloesch, et~al.
\newblock Anymal-a highly mobile and dynamic quadrupedal robot.
\newblock In {\em 2016 IEEE/RSJ International Conference on Intelligent Robots
  and Systems (IROS)}, pages 38--44. IEEE, 2016.

\bibitem{jenelten2021TAMOLS}
Fabian Jenelten, Ruben Grandia, Farbod Farshidian, and Marco Hutter.
\newblock Tamols: Terrain-aware motion optimization for legged systems.
\newblock {\em submitted to IEEE Transactions on Robotics}, 2021.

\bibitem{jenelten2020perceptive}
Fabian Jenelten, Takahiro Miki, Aravind~E Vijayan, Marko Bjelonic, and Marco
  Hutter.
\newblock Perceptive locomotion in rough terrain--online foothold optimization.
\newblock {\em IEEE Robotics and Automation Letters}, 5(4):5370--5376, 2020.

\bibitem{khattak2020complementary}
Shehryar Khattak, Huan Nguyen, Frank Mascarich, Tung Dang, and Kostas Alexis.
\newblock Complementary multi--modal sensor fusion for resilient robot pose
  estimation in subterranean environments.
\newblock In {\em 2020 International Conference on Unmanned Aircraft Systems
  (ICUAS)}, pages 1024--1029. IEEE, 2020.

\bibitem{kim2020vision}
D~Kim, D~Carballo, J~Di~Carlo, B~Katz, G~Bledt, B~Lim, and Sangbae Kim.
\newblock Vision aided dynamic exploration of unstructured terrain with a
  small-scale quadruped robot.
\newblock In {\em 2020 IEEE International Conference on Robotics and Automation
  (ICRA)}, pages 2464--2470. IEEE, 2020.

\bibitem{kolter2009stereo}
J~Zico Kolter, Youngjun Kim, and Andrew~Y Ng.
\newblock Stereo vision and terrain modeling for quadruped robots.
\newblock In {\em 2009 IEEE International Conference on Robotics and
  Automation}, pages 1557--1564. IEEE, 2009.

\bibitem{kulkarni2021autonomous}
Mihir Kulkarni, Mihir Dharmadhikari, Marco Tranzatto, Samuel Zimmermann, Victor
  Reijgwart, Paolo~De Petris, Huan Nguyen, Nikhil Khedekar, Christos
  Papachristos, Lionel Ott, Roland Siegwart, Marco Hutter, and Kostas Alexis.
\newblock Autonomous teamed exploration of subterranean environments using
  legged and aerial robots.
\newblock In {\em 2022 IEEE International Conference on Robotics and Automation
  (ICRA)}, Philadelphia (PA), USA, 2022. IEEE.

\bibitem{lynen13robust}
S~Lynen, M~Achtelik, S~Weiss, M~Chli, and R~Siegwart.
\newblock A robust and modular multi-sensor fusion approach applied to mav
  navigation.
\newblock In {\em Proc. of the IEEE/RSJ Conference on Intelligent Robots and
  Systems (IROS)}, 2013.

\bibitem{ma2022combining}
Yuntao Ma, Farbod Farshidian, Takahiro Miki, Joonho Lee, and Marco Hutter.
\newblock Combining learning-based locomotion policy with model-based
  manipulation for legged mobile manipulators.
\newblock {\em IEEE Robotics and Automation Letters}, 7(2):2377--2384, 2022.

\bibitem{wild_anymal}
Takahiro Miki, Joonho Lee, Jemin Hwangbo, Lorenz Wellhausen, Vladlen Koltun,
  and Marco Hutter.
\newblock Learning robust perceptive locomotion for quadrupedal robots in the
  wild.
\newblock {\em Science Robotics}, 7(62):eabk2822, 2022.

\bibitem{nguyen2012modeling}
Chuong~V. Nguyen, Shahram Izadi, and David Lovell.
\newblock Modeling kinect sensor noise for improved 3d reconstruction and
  tracking.
\newblock In {\em 2012 Second International Conference on 3D Imaging, Modeling,
  Processing, Visualization Transmission}, pages 524--530, 2012.

\bibitem{nishino2017cupy}
ROYUD Nishino and Shohei Hido~Crissman Loomis.
\newblock Cupy: A numpy-compatible library for nvidia gpu calculations.
\newblock {\em 31st confernce on neural information processing systems}, 151,
  2017.

\bibitem{oleynikova2017voxblox}
Helen Oleynikova, Zachary Taylor, Marius Fehr, Roland Siegwart, and Juan Nieto.
\newblock Voxblox: Incremental 3d euclidean signed distance fields for on-board
  mav planning.
\newblock In {\em IEEE/RSJ International Conference on Intelligent Robots and
  Systems (IROS)}, 2017.

\bibitem{overbye2021gvom}
Timothy Overbye and Srikanth Saripalli.
\newblock G-vom: A gpu accelerated voxel off-road mapping system, 2021.

\bibitem{pan2021}
Yiyuan Pan, Xuecheng Xu, Xiaqing Ding, Shoudong Huang, Yue Wang, and Rong
  Xiong.
\newblock Gem: Online globally consistent dense elevation mapping for
  unstructured terrain.
\newblock {\em IEEE Transactions on Instrumentation and Measurement}, 70:1--13,
  2021.

\bibitem{NEURIPS2019_9015}
Adam Paszke, Sam Gross, Francisco Massa, Adam Lerer, James Bradbury, Gregory
  Chanan, Trevor Killeen, Zeming Lin, Natalia Gimelshein, Luca Antiga, Alban
  Desmaison, Andreas Kopf, Edward Yang, Zachary DeVito, Martin Raison, Alykhan
  Tejani, Sasank Chilamkurthy, Benoit Steiner, Lu~Fang, Junjie Bai, and Soumith
  Chintala.
\newblock Pytorch: An imperative style, high-performance deep learning library.
\newblock In H.~Wallach, H.~Larochelle, A.~Beygelzimer, F.~d\textquotesingle
  Alch\'{e}-Buc, E.~Fox, and R.~Garnett, editors, {\em Advances in Neural
  Information Processing Systems 32}, pages 8024--8035. Curran Associates,
  Inc., 2019.

\bibitem{quigley2009ros}
Morgan Quigley, Ken Conley, Brian Gerkey, Josh Faust, Tully Foote, Jeremy
  Leibs, Rob Wheeler, Andrew~Y Ng, et~al.
\newblock Ros: an open-source robot operating system.
\newblock In {\em ICRA workshop on open source software}, volume~3, page~5.
  Kobe, Japan, 2009.

\bibitem{rudin2022learning}
Nikita Rudin, David Hoeller, Philipp Reist, and Marco Hutter.
\newblock Learning to walk in minutes using massively parallel deep
  reinforcement learning.
\newblock In {\em Conference on Robot Learning}, pages 91--100. PMLR, 2022.

\bibitem{Rusu_ICRA2011_PCL}
Radu~Bogdan Rusu and Steve Cousins.
\newblock {3D is here: Point Cloud Library (PCL)}.
\newblock In {\em {IEEE International Conference on Robotics and Automation
  (ICRA)}}, Shanghai, China, May 9-13 2011.

\bibitem{stoelzle2022reconstructing}
Maximilian Stölzle, Takahiro Miki, Levin Gerdes, Martin Azkarate, and Marco
  Hutter.
\newblock Reconstructing occluded elevation information in terrain maps with
  self-supervised learning.
\newblock {\em IEEE Robotics and Automation Letters}, 7(2):1697--1704, 2022.

\bibitem{thrun2002probabilistic}
Sebastian Thrun.
\newblock Probabilistic robotics.
\newblock {\em Communications of the ACM}, 45(3):52--57, 2002.

\bibitem{tranzatto2021cerberus}
Marco Tranzatto, Frank Mascarich, Lukas Bernreiter, Carolina Godinho, Marco
  Camurri, Shehryar Masaud~Khan Khattak, Tung Dang, Victor Reijgwart, Johannes
  Loeje, David Wisth, et~al.
\newblock Cerberus: Autonomous legged and aerial robotic exploration in the
  tunnel and urban circuits of the darpa subterranean challenge.
\newblock {\em Journal of Field Robotics}, 2021.

\bibitem{rudolph2006multi}
Rudolph Triebel, Patrick Pfaff, and Wolfram Burgard.
\newblock Multi-level surface maps for outdoor terrain mapping and loop
  closing.
\newblock In {\em 2006 IEEE/RSJ International Conference on Intelligent Robots
  and Systems}, pages 2276--2282, 2006.

\bibitem{magana2019fast}
Octavio~Antonio Villarreal{-}Maga{\~{n}}a, Victor Barasuol, Marco Camurri,
  Michele Focchi, Luca Franceschi, Massimiliano Pontil, Darwin~G. Caldwell, and
  Claudio Semini.
\newblock Fast and continuous foothold adaptation for dynamic locomotion
  through cnns.
\newblock {\em IEEE Robotics and Automation Letters}, 4(2):2140--2147, 2019.

\bibitem{artplanner}
Lorenz Wellhausen and Marco Hutter.
\newblock Rough terrain navigation for legged robots using reachability
  planning and template learning.
\newblock In {\em 2021 IEEE/RSJ International Conference on Intelligent Robots
  and Systems (IROS)}, pages 6914--6921, 2021.

\bibitem{wermelinger2016}
Martin Wermelinger, Péter Fankhauser, Remo Diethelm, Philipp Krüsi, Roland
  Siegwart, and Marco Hutter.
\newblock Navigation planning for legged robots in challenging terrain.
\newblock In {\em 2016 IEEE/RSJ International Conference on Intelligent Robots
  and Systems (IROS)}, pages 1184--1189, 2016.

\bibitem{yangbowen2021}
Bowen Yang, Lorenz Wellhausen, Takahiro Miki, Ming Liu, and Marco Hutter.
\newblock Real-time optimal navigation planning using learned motion costs.
\newblock In {\em 2021 IEEE International Conference on Robotics and Automation
  (ICRA)}, pages 9283--9289, 2021.

\end{thebibliography}

\end{document}